\documentclass[letterpaper, 10 pt, conference]{ieeeconf}  

\IEEEoverridecommandlockouts                              
\usepackage{amsmath,amssymb,amsfonts}
\usepackage{booktabs}
\usepackage{multirow}
\usepackage{graphicx}
\usepackage{xcolor}
\usepackage{hyperref}
\usepackage{cite}
\usepackage{siunitx}
\usepackage{algorithm}
\usepackage{algorithmic}
\usepackage{amsthm}
\usepackage{comment}

\newcommand{\Jeff}{\mathbf{J}_\mathrm{eff}}
\newcommand{\Jtip}{\mathbf{J}_\mathrm{tip}}
\newcommand{\Jrcm}{\mathbf{J}_\mathrm{rcm}}
\newcommand{\Nrcm}{\mathbf{N}_\mathrm{rcm}}
\newcommand{\prcm}{\mathbf{p}_\mathrm{rcm}}
\newcommand{\ptip}{\mathbf{p}_\mathrm{tip}}
\newcommand{\smin}{\sigma_{\min}}
\newcommand{\vmin}{\hat{\mathbf{v}}_{\min}}
\newcommand{\qstar}{\mathbf{q}^{\star}}

\title{High-Precision RCM Motion for Ophthalmic Robotics via Hierarchical QP Control and Authority-Guided Recovery Planning}

\newif\ifanonymous
\anonymousfalse

\ifanonymous
\author{Anonymous Author(s)}
\else
\author{Huanyu Tian$^{1}$, Adriana Namour$^{1}$, Lingyun Zeng$^{1}$, Zicong Wu$^{1}$, Sebastien Ourselin$^{1}$, \\ Lyndon da Cruz$^{2,\star}$ and Christos Bergeles$^{1,\star}$
\thanks{*This work was supported by the National Institutes for Health and Care Research (NIHR506200), the EPSRC CDT in Smart Medical Imaging (EP/S022104/1), Moorfields Eye Charity through funding from the Sir Michael Uren, and Sir Joseph Hotung, Foundations. This work was also supported by the Bota Systems Sponsoring Program for Physical AI Research and the NVIDIA Academic Grant Program. The views expressed are those of the author(s) and not necessarily those of the National Health Service, the National Institute for Health Research, or the Department of Health and Social Care. For the purpose of open access, the author has applied a CC BY public copyright license to any Author Accepted Manuscript version arising from this submission.}
\thanks{$^{1}$H.~Tian, A.~Namour, L.~Zeng, Z.~Wu, S. Ourselin and C.~Bergeles are with the School of Biomedical Engineering \& Imaging Sciences, King's College London, UK. Corresponding author: {\tt\small huanyu.tian@kcl.ac.uk}}
\thanks{$^{2}$L.~da Cruz is with Moorfields Eye Hospital, NHS Foundation Trust, London, UK.}
\thanks{$^\star$ denotes equal contribution.}
}
\fi

\begin{document}
\maketitle

\begin{abstract}
General-purpose manipulators can position ophthalmic instruments under a software Remote Centre of Motion (RCM). However, at some reachable configurations that are nonsingular for the tip task alone, the RCM and tip-tracking objectives conflict and the servo oscillates.
We formalise this failure as a constraint-induced task singularity: the RCM-projected tip Jacobian loses rank while the unprojected tip Jacobian stays well conditioned. Thus, the effective tip authority collapses although the target remains reachable.
We present a high-precision RCM servo architecture with built-in singularity recovery: for each commanded target, an authority-guided planner selects a well-conditioned arm configuration that reaches it and plans an RCM-consistent route, executed by a fixed hierarchical-QP controller.
On a 200-scenario kinematic benchmark, the complete nominal-and-recovery protocol reaches 128/128 near-singular targets and 71/72 near-joint-limit targets (100\% and 99\%, respectively), with the software-RCM residual below 0.142 mm.
On a physical 6-DOF arm it executes across four RCM entry points, with a worst-case kinematically reconstructed RCM residual of 178 $\mu m$ and a median tip-regulation error of 32 $\mu m$, within the 200 $\mu m$ control requirement adopted for the ophthalmic task.
\end{abstract}


\section{Introduction}
\begin{figure}[!t]
\centering
\includegraphics[width=0.95\linewidth]{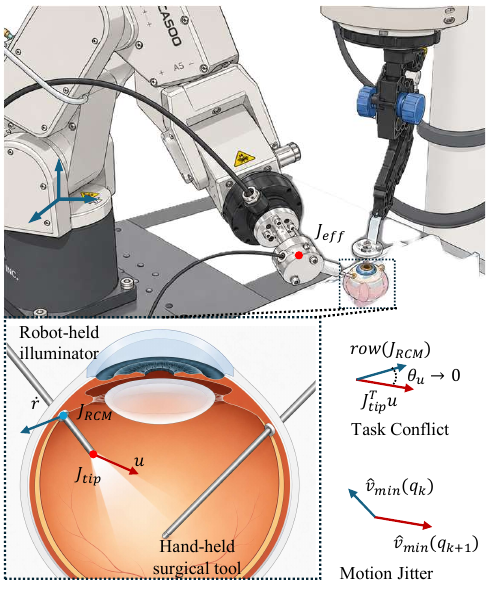}
\caption{\textbf{Setup and failure mechanism.}
A robot-held illuminator is servoed to a target under a software RCM at its scleral entry; the hand-held surgical tool is the surgeon's and out of scope.
\emph{Motion Conflict}: the tip-progress gradient aligns with the RCM constraint directions, so the effective tip authority $\smin(\Jeff)$ collapses while the unprojected tip Jacobian stays well conditioned, a \emph{constraint-induced task singularity}.
\emph{Motion Jitter}: oscillatory trajectories exhibit rapid rotation of the weak joint-space direction and large step-to-step changes in the amplified command near the authority collapse (Sec.~\ref{sec:mechanism}).}
\label{fig:mechanism}
\end{figure}
Robot-assisted vitreoretinal surgery is a keyhole task: a slender instrument is steered inside the eye while its shaft pivots about a fixed scleral entry, the Remote Centre of Motion (RCM)~\cite{VANDERPOORTEN2026751}.
In vitreoretinal surgery, both the robot tip error and the RCM residual must be held to maximum \SI{200}{\micro\meter}, against retinal structures only a few hundred micrometres thick, leaving little margin to trade between the two objectives~\cite{edwards2018first,gijbels2016towards}.
Purpose-built platforms realise the RCM through a mechanical linkage~\cite{mamplekos2022modelling,zhang2025deep,bian2026autonomous,smits2020synthesis,chen2018intraocular}, which may also be adaptable \cite{inagaki2024analyzing,xia2024online}. A general-purpose manipulator, e.g. such as that used in \cite{posselli2025head}, however, enforces it through software and control, and the constraint becomes coupled with tip tracking and competes for the same joint velocities.

At certain reachable configurations that are nonsingular for the unconstrained tip-position task, this coupling becomes a \emph{constraint-induced task singularity}: the joint motion that advances the tip is the same motion that sweeps the shaft laterally through the sclera, i.e.\ the tip-progress gradient aligns with the RCM constraint directions. In such a case, no RCM-consistent joint velocity produces the required first-order tip progress (Fig.~\ref{fig:mechanism}).
Partial alignment can be accommodated within the residual null space by a prioritised controller~\cite{Yang2026TMRB,Yan2022ICRA}. As the alignment approaches completion, however, the available first-order tip authority vanishes. 
A loose RCM tolerance may absorb the conflict through temporary constraint violation, whereas the tight ophthalmic tolerance leads to oscillatory tip motion or loss of tip progress toward the target (Sec.~\ref{sec:mechanism}).


In prior work, programmable RCM enforcement is realised with task-priority (hierarchical) control and solved by hierarchical quadratic programming (HQP). The RCM residual is imposed as the high-priority level, and tip tracking is resolved on the remaining subspace. The constraint is written as an equality in the hierarchy or as an inequality in a positional- or velocity-level QP~\cite{li2026rcm,koyama2022autonomous,su2021toward,marinho2019dynamic,hu2020hierarchical}. The residual null space is used for a  redundancy resolution, typically toward a manipulability/conditioning measure~\cite{Aghakhani2013,hu2020hierarchical}.
Because this resolution is local and provides no path-level lookahead, it cannot route around a singularity lying on the nominal path. The underlying projected-Jacobian rank loss is known in task-priority control as an \emph{algorithmic singularity}~\cite{escande2014hierarchical}.

Two gaps therefore remain. First, software-RCM work reports RCM and tip accuracy, but rarely analyses the closed-loop oscillation produced by constrained-induced task singularity under a strict tolerance. Second, redundancy-resolution studies typically involve more redundant degrees of freedom and looser task tolerances, so the constrained null space retains enough dimensions, and the solver enough slack, to absorb the conflict, which then never appears as a task singularity.

We study a representative challenging case: a $6$-DOF arm performing 3D point positioning under a high-priority RCM constraint with bounded slack, where a single redundant degree of freedom (roll about the shaft) leaves little room to reconcile the RCM and tip objectives once joint limits are active.
The contributions of this work are:
\begin{enumerate}
\item
A bounded-slack HQP controller that enforces the RCM while servoing the tip.
\item
An authority-guided recovery planner, built on our characterisation of the constraint-induced task singularity, that acts only on the reference the controller tracks.
\end{enumerate}
These are evaluated on a $200$-scenario kinematic benchmark of rare worst-case configurations and a collapse case study, and demonstrated in tip-positioning experiments on a robotic arm within the task's \SI{200}{\micro\meter} control requirement.

\section{Problem Setup and Task-Singularity Characterisation}
\label{sec:setup}
We consider a $6$-DOF manipulator with joint configuration $\mathbf{q}\in\mathbb{R}^{6}$ holding a rigid instrument, with tip position $\ptip(\mathbf{q})\in\mathbb{R}^3$ and tip Jacobian $\Jtip(\mathbf{q})=\partial\ptip/\partial \mathbf{q}\in\mathbb{R}^{3\times6}$. The joints obey box limits $\mathbf{q}_{\min}\le\mathbf{q}\le\mathbf{q}_{\max}$, with range centre $\mathbf{q}_c=\tfrac12(\mathbf{q}_{\min}+\mathbf{q}_{\max})$ and joint-limit margin $m(\mathbf{q})=\min_j\min(\mathbf{q}_j-\mathbf{q}_{\min,j},\,\mathbf{q}_{\max,j}-\mathbf{q}_j)$.
A fixed RCM point $\prcm\in\mathbb{R}^3$ marks the scleral entry (Fig.\ \ref{fig:mechanism}).
The task is to drive $\ptip$ to a target $\mathbf{x}_g$ inside a workspace cone about $\prcm$ while keeping the instrument shaft passing through $\prcm$\footnote{Entry-point placement is a pre-operative kinematic choice: a suitable entry places the intended targets well inside the RCM-feasible cone, so reachability is a solved step. The failures we study are therefore failures of \emph{executability}, not of reachability. In these failure cases, under the tight bounded-slack RCM constraint and a finite time limit, the servo cannot smoothly reach a target that IK can solve.}.

\subsection{Low-level RCM QP}
\label{sec:qp}

All reference-generation policies drive the same velocity-level QP. Its bounded-slack lexicographic structure prioritises the RCM residual over tip tracking while retaining feasibility when the objectives conflict~\cite{li2026rcm,hu2020hierarchical}.

Let $\mathbf{a}(\mathbf{q})$ be the instrument-axis direction and $\mathbf{p}_{\mathrm{shaft}}(\mathbf{q})$ a reference point on that axis, both given by the robot forward kinematics. 
The shaft-to-pivot residual $\mathbf{r}(\mathbf{q})\in\mathbb{R}^{2}$ consists of the two lateral components of $(\mathbf{I}-\mathbf{a}\mathbf{a}^{\top})(\prcm-\mathbf{p}_{\mathrm{shaft}})$; the axial component along the shaft is unconstrained. Its Jacobian is $\Jrcm(\mathbf{q})=\partial\mathbf{r}/\partial\mathbf{q}\in\mathbb{R}^{2\times6}$.
Let $\mathbf{e}(\mathbf{q})=\mathbf{x}^{\mathrm{cmd}}-\ptip(\mathbf{q})$ be the commanded tip error, where $\mathbf{x}^{\mathrm{cmd}}$ is the Cartesian reference input to the controller; generating this reference is the subject of Sec.~\ref{sec:method}.

The QP is \emph{lexicographic} in two levels, with a \emph{bounded-slack} RCM constraint. Rather than driving the residual exactly to zero, the first level admits a normalised slack $\xi\in[0,1]$ that measures how far the residual is pushed past a hard tolerance band $\tau$ and minimises that slack. The second level, on the first level's optimal set, tracks the tip while regularising configuration,
\begin{subequations}
\begin{align}
\textbf{L1:}\ \ & \min_{\dot{\mathbf{q}},\,\xi}\ w_1\,\xi + w_2\,\xi^2 \label{eq:qp_l1}\\
& \text{s.t.}\ \ \|\mathbf{r}(\mathbf{q})+h\,\Jrcm\dot{\mathbf{q}}\|_\infty\le \tau + \xi\,\bar s, \notag\\
& \hphantom{\text{s.t.}\ \ }\|\dot{\mathbf{q}}\|_\infty\le \dot{\mathbf{q}}_{\max}, \notag\\ & \hphantom{\text{s.t.}\ \ }0\le\xi\le1, \notag\\[3pt]
\textbf{L2:}\ \ &
\min_{(\dot{\mathbf{q}},\xi)\in\mathcal{S}_1}
\|\Jtip(\mathbf{q})\dot{\mathbf{q}}-\kappa\mathbf{e}(\mathbf{q})\|^2
+\mu\|\dot{\mathbf{q}}-\dot{\mathbf{q}}_{\mathrm{reg}}\|^2,
\label{eq:qp_l2}
\end{align}
\end{subequations}
where $h$ is the servo period, $w_1,w_2>0$ penalise the slack, $\dot{\mathbf{q}}_{\max}$ bounds the joint velocity, $\bar s$ scales the normalised slack so that the residual may exceed $\tau$ by at most $\bar s$, $\mathcal{S}_1$ is the set of L1-optimal pairs
$(\dot{\mathbf{q}},\xi)$, allowing a small numerical margin
from the optimum $\xi^\star$, $\kappa$ is the tip servo gain, and $\dot{\mathbf{q}}_{\mathrm{reg}}$, weighted by $\mu$, biases the solution toward a nominal joint velocity. Like $\mathbf{x}^{\mathrm{cmd}}$, $\dot{\mathbf{q}}_{\mathrm{reg}}$ is supplied by the reference-generation policy (Sec.~\ref{sec:gating}). The $\ell_\infty$ residual bound splits into two linear inequalities per component, so L1 remains a QP solvable at the servo rate.

Solving L1 \eqref{eq:qp_l1} confines the L2 motion \eqref{eq:qp_l2}, to first order, to the RCM null space, which admits the projector
\begin{equation}
\Nrcm(\mathbf{q}) = \mathbf{I} - \Jrcm(\mathbf{q})^{\dagger}\Jrcm(\mathbf{q}),
\label{eq:nrcm}
\end{equation}
so any RCM-consistent instantaneous motion realises tip velocity only through the \emph{effective} tip Jacobian
\begin{equation}
\Jeff(\mathbf{q}) = \Jtip(\mathbf{q})\,\Nrcm(\mathbf{q}).
\label{eq:jeff}
\end{equation}

Throughout, $\smin(\cdot)$ and $\sigma_{\max}(\cdot)$ denote the smallest and largest singular values of a matrix. We call $\smin(\Jeff)$ the \emph{effective (tip) authority}, i.e.\ the minimum singular value of the RCM-projected tip Jacobian, the tip speed per unit joint speed along the tip's least controllable direction (units \si{\metre\per\radian}; all singular values in this paper carry this unit), and $\mathrm{cond}(\Jeff)=\sigma_{\max}(\Jeff)/\smin(\Jeff)$ the \emph{conditioning}. We write $\mathbf{N}^{\mathrm{tip}}_{\mathrm{rcm}}(\mathbf{q})=\mathbf{I}-\mathbf{J}_{\mathrm{aug}}^{\dagger}\mathbf{J}_{\mathrm{aug}}$, with the stacked Jacobian $\mathbf{J}_{\mathrm{aug}}=[\Jrcm^{\top}\ \ \Jtip^{\top}]^{\top}$, for the projector onto RCM-null-space motions that additionally leave the tip velocity unchanged (the instrument self-motion), used later for joint-limit centering that cannot disturb tip tracking.

As the admitted slack decreases, the feasible velocity set approaches $\ker(\Jrcm)$ and tip motion becomes increasingly governed by $\Jeff$, exposing the authority collapse analysed below. All methods use this fixed QP and differ only in their reference-generation policies\footnote{The effective-Jacobian quantities $\Jtip,\Jrcm,\Nrcm,\Jeff$ follow \eqref{eq:nrcm}--\eqref{eq:jeff} in both the run-time controller and the offline diagnostics. A small Tikhonov damping used by the servo near the collapse does not alter the qualitative $\smin(\Jeff)\!\to\!0$ behaviour.}.

\subsection{Task-singularity characterisation}
\label{sec:mechanism}
The failure is constraint-induced and configuration-dependent, and it manifests only when the admitted slack is tight.
We state it in terms the standard tools (SVD conditioning \cite{golub2013matrix}, singularity-robust IK \cite{xing2026inverse}, manipulability \cite{jaquier2021geometry}) already provide, and add a single new descriptor, the weak-direction rotation defined below.
The effective gain of $\Jeff$ along a unit tip direction $\mathbf{u}$ is
\begin{equation}
\|\Jeff^{\top}\mathbf{u}\| = \|\Nrcm\Jtip^{\top}\mathbf{u}\| = \|\Jtip^{\top}\mathbf{u}\|\,\sin\theta_{u},
\label{eq:tasksing-overlap}
\end{equation}
with $\theta_{u}$ the principal angle between the tip-error gradient $\Jtip^{\top}\mathbf{u}$ and the row space of $\Jrcm$.
When $\|\Jtip^{\top}\mathbf{u}\|$ stays bounded away from zero (the unprojected tip Jacobian is nonsingular) but $\sin\theta_{u}\to0$, the effective gain and $\smin(\Jeff)$ vanish: a rank loss of $\Jeff$, not of $\Jtip$, induced entirely by the RCM constraint, which is the constraint-induced task singularity introduced above. We call this vanishing of the effective authority the \emph{authority collapse}, or simply the collapse.

Although this rank loss corresponds algebraically to an algorithmic singularity~\cite{escande2014hierarchical}, we focus on its trajectory-level manifestation under tight RCM tolerances and evaluate weak-direction rotation as an empirical indicator of oscillation. We then address the collapse through reference-level planning rather than controller damping (Sec.~\ref{sec:method}).

Let $\Jeff=\mathbf{U}\boldsymbol{\Sigma}\mathbf{V}_r^{\top}$ be the compact SVD, with $\boldsymbol{\Sigma}=\mathrm{diag}(\sigma_1,\sigma_2,\sigma_3)$, $\sigma_1\ge\sigma_2\ge\sigma_3\ge0$, and $\mathbf{V}_r\in\mathbb{R}^{6\times3}$; we set $\smin(\Jeff)=\sigma_3$ and denote its right-singular vector by $\vmin$, the joint-space direction associated with the least controllable tip-space mode (distinct from the exact self-motion null space of $\Jeff$).
Near rank loss, the bounded QP (Sec.~\ref{sec:qp}) becomes increasingly sensitive to this direction, and in the low-authority regime the solution is strongly activated along it.
We measure whether $\vmin$ rotates between consecutive servo steps, by the sign-aligned angle
\begin{equation}
\Delta\phi_k=\arccos\big|\vmin(\mathbf{q}_k)^{\top}\vmin(\mathbf{q}_{k+1})\big|\ \in[0^\circ,90^\circ],
\label{eq:vmin_rot}
\end{equation}
the absolute value removing the singular-vector sign ambiguity; $\vmin$ is interpreted only where $\smin(\Jeff)$ is sufficiently separated from the neighbouring singular value.
The mechanism we evaluate is that, in the low-authority regime, a rapidly rotating weak direction redirects the strongly activated joint-space command between steps and can reverse it, plausibly accounting for the observed command reversals and tip jitter, whereas a slowly rotating weak direction, even at small $\smin(\Jeff)$, redirects it little (Fig.~\ref{fig:mechanism}).
We therefore treat $\Delta\phi_k$ as an empirically supported indicator that distinguishes smooth from oscillatory low-authority trajectories (Sec.~\ref{sec:evidence}), rather than as a proven cause of oscillation.

\section{Authority-Guided Recovery Planning}
\label{sec:method}
\begin{figure}[!t]
\centering
\includegraphics[width=\linewidth]{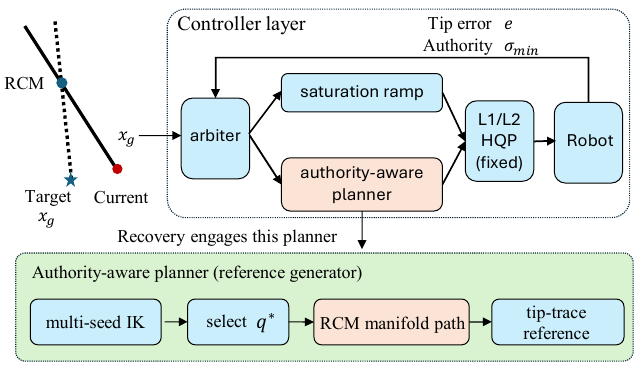}
\caption{\textbf{Unified authority-guided architecture.}
The upper layer supplies a saturated or planned reference and adds null-space joint centring; the fixed HQP executes the resulting command under the RCM constraint. When activated, the planner uses multi-seed IK to select a high-authority configuration and constructs an RCM-consistent path to it.}
\label{fig:system}
\end{figure}

The proposed architecture combines the fixed low-level QP with an upper layer that supplies saturated or planned references (Fig.~\ref{fig:system}). Effective authority governs planning activation and terminal-configuration selection, while null-space joint centring addresses proximity to joint limits. The planner does not impose a formal lower bound on authority along the generated path.

\subsection{Input-reference saturation (baseline policy)}
\label{sec:sat}
Direct servoing feeds the QP the full error $\mathbf{x}_g-\ptip(\mathbf{q}_k)$ every step, which near a collapse drives an aggressive descent into the ill-conditioned direction.
We instead ramp the commanded reference along the straight segment from the start tip $\mathbf{x}_0$ to the target $\mathbf{x}_g$,
\begin{subequations}\label{eq:sat}
\begin{align}
\mathbf{x}^{\mathrm{sat}}_k &= (1-s_k)\,\mathbf{x}_0 + s_k\,\mathbf{x}_g,\\
s_k &= \min\!\Big(\tfrac{k+1}{N_{\mathrm{path}}},\,1\Big),
\end{align}
\end{subequations}
where $N_{\mathrm{path}}$ limits the reference increment to \SI{0.5}{\milli\meter} per step and $\mathbf{x}^{\mathrm{cmd}}_k=\mathbf{x}^{\mathrm{sat}}_k$. This open-loop ramp reduces excitation of the weak direction but cannot avoid a low-authority region lying on the straight path.

\subsection{Authority-guided recovery via a well-conditioned configuration}
\label{sec:planner}
When the straight path crosses the collapse, reducing the reference increment is insufficient. The planner therefore selects a well-conditioned target-reaching configuration and constructs an RCM-consistent route to it in three steps.

\paragraph*{(i) Configuration selection}
The RCM constraint confines the instrument to the configuration manifold $\mathcal{M}=\{\mathbf{q}:\mathbf{r}(\mathbf{q})=\mathbf{0}\}$, on which the tip is actuated only through $\Jeff$, and $\Nrcm(\mathbf{q})$ projects onto the tangent space of $\mathcal{M}$ at $\mathbf{q}$. Among the configurations that reach $\mathbf{x}_g$ on $\mathcal{M}$ within tolerance and joint limits,
\begin{equation}
\begin{aligned}
\mathcal{Q}_g=\big\{\mathbf{q}:\ \|\mathbf{e}_{\mathrm{tip}}(\mathbf{q})\|\le\varepsilon_{\mathrm{tip}},\ \|\mathbf{r}(\mathbf{q})\|\le\varepsilon_{\mathrm{rcm}},\ \\\mathbf{q}_{\min}\le\mathbf{q}\le\mathbf{q}_{\max}\big\},
\label{eq:feasible_set}
\end{aligned}
\end{equation}
with tip residual $\mathbf{e}_{\mathrm{tip}}(\mathbf{q})=\ptip(\mathbf{q})-\mathbf{x}_g$ and acceptance tolerances $\varepsilon_{\mathrm{tip}},\varepsilon_{\mathrm{rcm}}$, a multi-seed constrained IK search returns a finite candidate set $\widehat{\mathcal{Q}}_g=\{\mathbf{q}^{(1)},\ldots,\mathbf{q}^{(N)}\}\subseteq\mathcal{Q}_g$, and we select the candidate with the highest effective authority,
\begin{equation}
\qstar=\arg\max_{\mathbf{q}\in\widehat{\mathcal{Q}}_g}\ \smin\!\big(\Jeff(\mathbf{q})\big),
\label{eq:configuration}
\end{equation}
selecting by feasibility first (only $\widehat{\mathcal{Q}}_g$ is considered) and effective authority second, so authority is never traded for feasibility.

\paragraph*{(ii) RCM-manifold path}
The start $\mathbf{q}_0$ is connected to $\qstar$ by a discrete flow of $K$ steps that stays on $\mathcal{M}$: at each increment a norm-capped tangent step advances toward $\qstar$ within $\ker\Jrcm$ \eqref{eq:proj_step}, and a Newton retraction $\Pi_{\mathrm{RCM}}$ (a Newton iteration on $\mathbf{r}(\mathbf{q})=\mathbf{0}$, terminating at $\|\mathbf{r}\|\le\varepsilon_{\mathrm{rcm}}$) corrects the second-order drift of the linear step \eqref{eq:reproject},
\begin{subequations}\label{eq:proj_step}
\begin{align}
\boldsymbol{\delta}_k &= \Nrcm(\mathbf{q}^{\mathrm{plan}}_k)\,k_p\big(\qstar - \mathbf{q}^{\mathrm{plan}}_k\big),\\
\dot{\mathbf{q}}_k &= \boldsymbol{\delta}_k \Big/ \underbrace{\max\!\big(1,\ \|\boldsymbol{\delta}_k\|/\dot{\mathbf{q}}_{\mathrm{cap}}\big)}_{\text{norm cap}}.
\end{align}
\end{subequations}
\begin{equation}
\mathbf{q}^{\mathrm{plan}}_{k+1} = \Pi_{\mathrm{RCM}}\big(\mathbf{q}^{\mathrm{plan}}_k+\Delta t\,\dot{\mathbf{q}}_k\big).
\label{eq:reproject}
\end{equation}
The resulting discrete path $\mathbf{q}^{\mathrm{plan}}_0=\mathbf{q}_0\!\to\!\cdots\!\to\!\mathbf{q}^{\mathrm{plan}}_K\!\approx\!\qstar$ consists of waypoints that satisfy the RCM tolerance by construction. During execution, the fixed HQP enforces the RCM constraint between successive references.
Although the fixed-tip self-motion set is generically one-dimensional for this $6$-DOF task, the RCM manifold traversed during recovery is generically four-dimensional. We use the projected flow because the terminal configuration is fixed by \eqref{eq:configuration} and the path must be generated online. Configuration selection maximises terminal authority, but the planner does not guarantee a lower bound on the path authority
\[
A_{\mathrm{path}}=\min_k\smin\!\big(\Jeff(\mathbf{q}^{\mathrm{plan}}_k)\big).
\]
Here, $k_p$ is the configuration-following gain, $\Delta t$ is the planning integration step, and $\dot{\mathbf{q}}_{\mathrm{cap}}$ limits the norm of each planned joint velocity. These parameters are fixed across all experiments.

\paragraph*{(iii) Reference exposure}
The planned joint path is exposed to the QP through a look-ahead Cartesian reference and an RCM-null-space configuration-guidance term. Let $k$ denote the current execution step, let $H\ge0$ be the fixed look-ahead horizon in waypoints, and define
\(
\ell_k=\min(k+H,K),
\)
where $K$ is the index of the final planned waypoint. The references are
\begin{subequations}\label{eq:lookahead}
\begin{align}
\mathbf{x}^{\mathrm{plan}}_k
&=
\ptip\!\left(\mathbf{q}^{\mathrm{plan}}_{\ell_k}\right),\\
\mathbf{g}^{\mathrm{plan}}_k
&=
\Nrcm(\mathbf{q}_k)\,
k_q\!\left(
\mathbf{q}^{\mathrm{plan}}_{\ell_k}-\mathbf{q}_k
\right),
\qquad k_q>0.
\end{align}
\end{subequations}
The Cartesian term $\mathbf{x}^{\mathrm{plan}}_k$ advances the tip reference to a future waypoint on the planned path. The joint-space term $\mathbf{g}^{\mathrm{plan}}_k$ biases the QP toward the corresponding planned configuration, with $\Nrcm(\mathbf{q}_k)$ restricting this guidance to the instantaneous RCM null space. Together, they help the controller remain on the intended configuration-space branch rather than tracking only the Cartesian tip trace.

\subsection{Unified authority-guided execution law}
\label{sec:gating}
The planner supplies the manifold reference $(\mathbf{x}^{\mathrm{plan}}_k,\mathbf{g}^{\mathrm{plan}}_k)$ of Sec.~\ref{sec:planner} \eqref{eq:lookahead} only for configurations flagged as high risk; elsewhere the direct goal, with joint-limit centering, suffices. We fold both regimes into one continuous reference fed to the low-level QP,
\begin{subequations}\label{eq:unified}
\begin{align}
\mathbf{x}^{\mathrm{cmd}}_k &= \alpha\,\mathbf{x}^{\mathrm{plan}}_k + (1-\alpha)\,\mathbf{x}^{\mathrm{sat}}_k,\\
\dot{\mathbf{q}}_{\mathrm{reg},k}&=\alpha\,\mathbf{g}^{\mathrm{plan}}_k+\beta\,\mathbf{N}^{\mathrm{tip}}_{\mathrm{rcm}}(\mathbf{q}_k)(\mathbf{q}_c-\mathbf{q}_k).
\end{align}
\end{subequations}
Here, $\alpha\in[0,1]$ weights the saturation and planned references, while $\beta\ge0$ adds null-space joint centring toward $\mathbf{q}_c$ without affecting the tip or RCM tasks. 
For the simulation comparisons within the proposed architecture, $\alpha=0$ denotes saturation execution and $\alpha=1$ plan following.

A manifold plan, and hence $\alpha=1$, is built only when the start configuration is flagged high-risk by fixed thresholds on its effective authority, conditioning, and joint-limit margin. The margin test matters because a configuration at a joint limit is well conditioned, so an authority-only gate would miss it. Otherwise no plan is built, $\alpha=0$, and the reference is the saturation ramp of Sec.~\ref{sec:sat}. If the tip later stalls short of the reach tolerance, a plan is rebuilt from the stuck configuration, so a start that the gate fails to flag is still recovered; this in-episode replanning is used in live execution, including the hardware study of Sec.~\ref{sec:hardware}, whereas the benchmark uses the scenario-level restart protocol of Sec.~\ref{sec:bench}.

In live execution, $\alpha$ is determined from the start-configuration authority $\smin(\Jeff(\mathbf{q}_0))$ and latched when the plan becomes available. In the simulation component comparisons, $\alpha\in\{0,1\}$ selects either direct control or plan following. Once planning is activated, it remains active for the rest of the execution so that the planned route is not abandoned after the authority recovers.
The centring weight $\beta$ rises linearly from $0$ to its maximum as the joint-limit margin $m(\mathbf{q}_k)$ falls between two fixed margins. Each episode is capped at a fixed number of steps, and all parameters of the law are held fixed across every experiment in Sec.~\ref{sec:experiments}.

\begin{algorithm}[!t]
\caption{Unified authority-guided execution law (per target).}
\label{alg:unified}
\begin{algorithmic}[1]
\REQUIRE target $\mathbf{x}_g$, start $\mathbf{q}_0$
\IF{the gate flags $\mathbf{q}_0$ as high-risk}
   \STATE build a plan to $\qstar$; set $\alpha=1$ in the benchmark, or compute $\alpha$ from the start authority in live execution; latch $\alpha$
\ELSE
   \STATE $\alpha\gets0$; use the saturation ramp (Sec.~\ref{sec:sat})
\ENDIF
\FOR{each servo step until reached, fault, or timeout}
   \STATE update $\beta$ from the joint-limit margin
   \STATE feed the reference of Eq.~\eqref{eq:unified} to the fixed QP
   \STATE if the tip stalls, rebuild the plan and re-latch $\alpha$
\ENDFOR
\STATE \textbf{benchmark supervisor:} if the target is not reached, restart once from $\mathbf{q}_0$ with closed-loop plan following (Sec.~\ref{sec:bench})
\end{algorithmic}
\end{algorithm}

\section{Experiments}
\label{sec:experiments}

The evaluation comprises a collapse case study, a fixed $200$-scenario kinematic benchmark, a recorded-configuration ablation, and a hardware execution study. The case study isolates the authority-collapse geometry and the main benchmark compares the complete proposed method with alternative configuration-management methods. The recorded-configuration ablation isolates the contribution of rerouting relative to input-reference saturation.

\subsection{Metrics and setup}
\label{sec:metrics}
All simulation runs use the kinematic model of the Mecademic Meca500 $6$-DOF arm (the hardware platform of Sec.~\ref{sec:hardware}) at a \SI{50}{\hertz} servo rate.
Tip jitter is the RMS of the second difference of the tip position sequence, in millimetres, and it captures step-to-step acceleration.
A target is reached if the tip error falls below a tolerance and stays there for a final window of $10$ servo steps. The reported steady-state error is the mean over this window. The tolerance is \SI{0.1}{\milli\meter} in the case study of Sec.~\ref{sec:experiments-collapse} and a relaxed \SI{0.15}{\milli\meter} in the larger, automatically evaluated benchmark of Sec.~\ref{sec:bench}.
For the mechanism analysis, we rank the benchmark scenarios by saturation-baseline jitter and compare the highest- and lowest-jitter $20$ scenarios. Ranking is independent of target reach.

\subsection{Empirical evidence for the mechanism}
\label{sec:evidence}
We compute the quantities of Sec.~\ref{sec:mechanism} from the logged trajectories of the oscillation and control groups.
In the highest-jitter scenarios $\smin(\Jeff)$ collapses to $10^{-8}$--$10^{-6}$ while $\Jtip$ stays well conditioned (median $\mathrm{cond}(\Jtip)=2.84$), consistent with a rank loss of $\Jeff$, rather than of $\Jtip$ \eqref{eq:tasksing-overlap}.
All four low-authority statistics separate the extreme-jitter groups strongly (AUC $0.95$--$0.98$): the minimum effective authority $\smin(\Jeff)$ at $0.98$, the maximum $\mathrm{cond}(\Jeff)$ at $0.97$, and the sign-aligned weak-direction rotation $\Delta\phi_k$ \eqref{eq:vmin_rot} at $0.95$/$0.96$ (max/p95), the descriptor that also localises \emph{when} the oscillation occurs.\footnote{The downstream signature is consistent: joint-jerk peaks co-locate with minima of $\smin(\Jeff)$ in $40\%$ of the oscillation group vs.\ $0\%$ of the control group, and on the deepest collapses the realised joint-velocity \emph{command} reverses by up to $179^\circ$ between steps, redirecting the amplified command almost completely.
This command reversal is distinct from, and larger than, the $\le\!90^\circ$ sign-aligned turn of the weak \emph{subspace} $\vmin$.}

\subsection{Case study: the authority-collapse geometry}
\label{sec:experiments-collapse}
\begin{figure}[!t]
\centering
\includegraphics[width=0.9\linewidth]{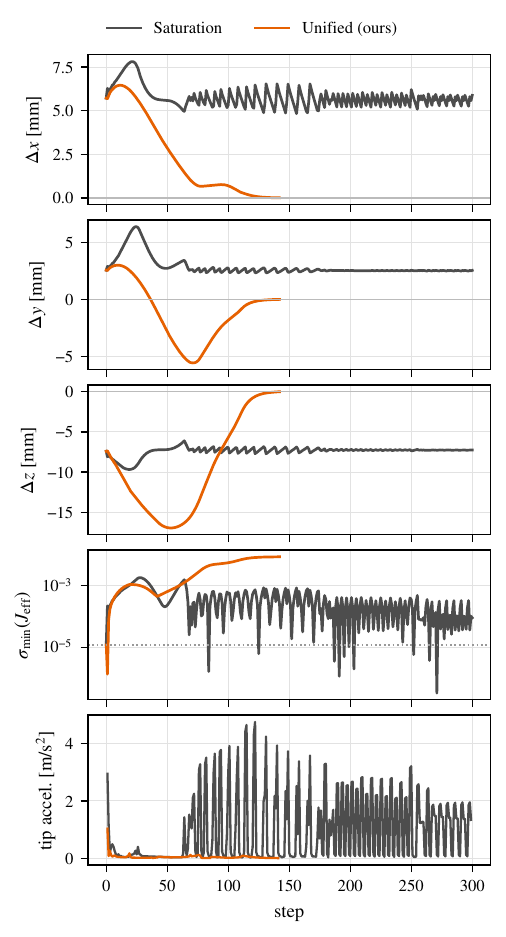}
\caption{\textbf{Collapse case study: planner versus input saturation.}
\textbf{(a)--(c)} Target-relative tip coordinates. The planner initially moves away from the target to bypass the authority collapse and then converges, whereas saturation stalls and oscillates.
\textbf{(d)} Effective authority $\smin(\Jeff)$ on a logarithmic scale. The planner maintains authority near $10^{-3}$, while saturation repeatedly enters the low-authority region; values below $10^{-5}$ are clipped at the dotted plotting floor. The low-authority intervals coincide with sustained jitter.
\textbf{(e)} Tip acceleration. Following an initial transient, the planner remains smooth, whereas saturation exhibits oscillations of up to $4$--$5$\,m/s$^2$.}
\label{fig:trajectory}
\end{figure}

The case study uses a single geometry in which the straight-line path from the start to the target crosses a deep authority collapse while the unprojected tip Jacobian remains well conditioned.
On this geometry, the authority-guided planner reaches the target under the tight \SI{0.1}{\milli\meter} tolerance by routing around the mid-path collapse, holding the maximum RCM residual at \SI{0.10}{\milli\meter}; input saturation, confined to the straight path, cannot clear the collapse and settles at a sustained offset (RCM residual \SI{0.12}{\milli\meter}). The routed solution is also more than an order of magnitude smoother, cutting tip jitter from \SI{1.05}{\milli\meter} to \SI{0.043}{\milli\meter}.
Logged authority values confirm the collapse is a loss of instantaneous controllability rather than of reachability: $\smin(\Jeff)$ reaches its minimum ($10^{-8}$--$10^{-6}$) mid-path while the terminal configuration recovers to $\approx1.5$--$1.7\times10^{-3}$, and a pre-operative multi-seed search finds alternative IK solutions with $\smin(\Jeff)>10^{-3}$: a well-conditioned configuration exists, and the nominal path to it is ill-conditioned.
The planner's selected configuration and planned path keep $\smin(\Jeff)$ near $10^{-3}$ throughout, so the weak direction never dominates the command. Fig.~\ref{fig:trajectory} shows the corresponding trajectories, authority profiles, and tip accelerations for both policies.

\subsection{Evaluation across an extended set of scenarios}
\label{sec:bench}
Beyond the single geometry of Sec.~\ref{sec:experiments-collapse}, we evaluate on a fixed benchmark of $200$ scenarios: $128$ \emph{singularity} scenarios whose starts are seeded inside the authority-collapse region, with $\smin(\Jeff)$ below a seeding threshold of $1.2\times10^{-5}$\,\si{\metre\per\radian} (deeper than the run-time gate threshold, which plays no role in seeding), and $72$ \emph{near-limit} scenarios that additionally stress configurations close to a joint limit.
Scenarios are generated with fixed seeds, so every run is reproducible.
 
\paragraph*{Procedure}
We compare the proposed method with two local configuration-management baselines~\cite{Aghakhani2013,hu2020hierarchical}: a QP \emph{configuration attractor} toward the joint-range centre $\mathbf{q}_c$, and RCM \emph{null-space resolution} toward an authority-optimised reference configuration $\mathbf{q}_{\mathrm{ref}}$. For each target, $\mathbf{q}_{\mathrm{ref}}$ is selected using the same feasible candidate set and authority criterion as $\qstar$ in Eq.~\eqref{eq:configuration}. The null-space baseline applies an instantaneous projected motion toward this configuration, whereas the proposed method constructs and follows an RCM-consistent path to it.  All methods use the same low-level QP, scenarios, and \SI{0.15}{\milli\meter} reach tolerance.
 
Under the pre-specified benchmark protocol, the proposed method consists of a nominal execution followed, if necessary, by one recovery execution. For a low-risk start, the nominal execution uses the saturation ramp; for a gate-flagged start, it uses open-loop plan following. If the nominal execution does not reach the target within the step limit, the scenario is restarted once from the same initial configuration using closed-loop plan following.
Input-reference saturation and plan following are compared separately as components of the proposed method in the controlled case study of Sec.~\ref{sec:experiments-collapse} and in the recorded-configuration ablation below.
 
\begin{table*}[!t]
\centering
\caption{Fixed $200$-scenario benchmark ($128$ singularity $+\,72$ near-limit),
split by failure axis: reach rate at the \SI{0.15}{\milli\meter} tolerance, final tip error (median and $95$th percentile), and worst-case RCM residual.}
\label{tab:bench-agg}
\small
\begin{tabular}{l cccc cccc}
\toprule
& \multicolumn{4}{c}{near-limit ($n{=}72$)} & \multicolumn{4}{c}{singularity ($n{=}128$)}\\
\cmidrule(lr){2-5}\cmidrule(lr){6-9}
policy & reach & tip med & tip p95 & RCM max & reach & tip med & tip p95 & RCM max\\
       & (\%)  & (mm)    & (mm)    & (mm)    & (\%)  & (mm)    & (mm)    & (mm)\\
\midrule
Configuration attractor       & 62 & 0.134 & 0.68 & 0.142 & 77 & 0.089 & 0.42 & 0.127\\
Null-space resolution   & 6  & 1.68  & 13.8 & 0.142 & 19 & 0.990 & 10.1 & 0.142\\
Unified method (ours)    & \textbf{99} & 0.075 & 0.10 & 0.142 & \textbf{100} & 0.099 & 0.12 & 0.142\\
\bottomrule
\end{tabular}
\end{table*}
 
Table~\ref{tab:bench-agg} reports the reach rate, final tip error, and RCM residual for the two scenario groups.
The configuration attractor reaches \SI{77}{\percent} of the singularity scenarios and \SI{62}{\percent} of the near-limit scenarios, whereas null-space resolution reaches only \SI{19}{\percent} and \SI{6}{\percent}, respectively; both exhibit multi-millimetre tip-error tails.

The complete nominal-and-recovery protocol reaches \SI{100}{\percent} of the singularity scenarios and \SI{99}{\percent} of the near-limit scenarios, with a worst-case RCM residual of \SI{0.142}{\milli\meter} and a p95 final tip error no greater than \SI{0.12}{\milli\meter}. Unlike the single-execution baselines, it permits one recovery execution after a failed nominal attempt.  For a deterministic baseline, restarting from the same initial configuration with unchanged inputs and parameters reproduces the same trajectory; the recovery execution instead changes the reference-generation policy to escape the stalled configuration.
These results show that the complete nominal-and-recovery method is substantially more reliable than instantaneous configuration-management alternatives on the seeded failure configurations.

 
One near-limit scenario remains unreached: neither the nominal execution nor the permitted recovery execution of the unified method reaches the target within the prescribed step limits. None of the evaluated methods converges to within \SI{5}{\milli\meter} of the target on this scenario.
\paragraph*{Recorded-configuration ablation}
To isolate the contribution of rerouting from input-reference saturation on configurations not designed from the authority criterion, we replay in simulation $28$ further tip-positioning problems whose start, target and RCM point are taken from poses recorded on the physical platform at eight scleral entry points, each start projected onto the RCM manifold so that it begins inside the constraint band.\footnote{The episode step budget is doubled for this set, whose tip travel (median \SI{17.6}{\milli\meter}) is about twice that of the seeded scenarios (median \SI{8.3}{\milli\meter}); all parameters of the execution law are unchanged. Raising the budget does not alter the saturation result ($11/28$ at either budget).}
Given a single execution per policy, input-reference saturation reaches $11/28$ of these targets with a median final tip error of \SI{3.34}{\milli\meter}, whereas plan following reaches $25/28$ with a median of \SI{0.083}{\milli\meter}; the software-RCM residual stays below \SI{0.142}{\milli\meter} for both. Plan following reaches $16$ targets that saturation does not, and saturation two that it does not. Slowing the reference along the direct route therefore leaves most of these targets unreached, whereas routing to a well-conditioned configuration reaches all but three.
\subsection{Hardware integration and real-time execution}
\label{sec:hardware}
The kinematic benchmark evaluates singularity recovery, whereas the hardware study evaluates real-time execution under a finite control rate, communication delay, and physical velocity tracking. The hardware experiment therefore tests physical execution of the architecture rather than independently comparing singularity-recovery performance.


\paragraph*{Platform} A $6$-DOF Mecademic Meca500 arm is commanded in joint-velocity mode over TCP/IP (\texttt{mecademicpy}). A Bota Systems MiniONE Pro force/torque sensor is used only for hand guiding before RCM point registration. At registration, $\prcm$ is fixed in the robot base frame from the calibrated robot configuration and tool geometry. During tracking, the tip position and RCM residual are reconstructed from forward kinematics. The force/torque sensor does not update either quantity. 
These kinematically reconstructed quantities directly evaluate regulation of the tip and software-RCM tasks defined by the controller, rather than absolute positioning accuracy in an independently measured anatomical frame.
The same QP formulation, constraint tolerances, and dimensionless policy parameters are used as in simulation, with time-discretisation terms evaluated at \SI{50}{\hertz}.

\paragraph*{Procedure} With the RCM entry point fixed, an operator issues
3D tip targets by teleoperation. Each is servoed by the fixed QP under the unified law (Sec.~\ref{sec:planner}--\ref{sec:gating}), and we log the realised tip trajectory, joint command, and RCM residual at the control rate using the definitions of Sec.~\ref{sec:metrics}. Results per entry point and pooled are given in Table~\ref{tab:hardware}. We report the reach rate at the \SI{0.15}{\milli\meter} tolerance, median final tip-regulation error, median and worst-case kinematically reconstructed RCM residual, and median tip jitter under the unified law. Metrics are evaluated over the RCM-constrained tracking phase and the pre-engagement approach before the admittance RCM lock is excluded.

\begin{table}[!t]
\centering
\caption{\textbf{Hardware test} across four RCM entry points and
$47$ teleoperated 3D targets. Results are reported per entry point
and pooled.}
\label{tab:hardware}
\small
\begin{tabular}{lccccc}
\toprule
RCM & \#tgt & reach & tip med. & RCM med./max & jitter\\
 & & (\SI{0.15}{\milli\meter}) & (mm) & (mm) & (mm)\\
\midrule
P1 & 8 & 8/8 & 0.073 & 0.007 / 0.178 & 0.016\\
P2 & 10 & 10/10 & 0.064 & 0.005 / 0.103 & 0.023\\
P3 & 19 & 16/19 & 0.026 & 0.007 / 0.169 & 0.021\\
P4 & 10 & 10/10 & 0.022 & 0.004 / 0.098 & 0.023\\
\midrule
 & 47 & 44\,(\SI{94}{\percent}) & 0.032 & 0.006 / 0.178 & 0.022\\
\bottomrule
\end{tabular}
\end{table}

\begin{figure}[!t]
\centering
\includegraphics[width=\linewidth]{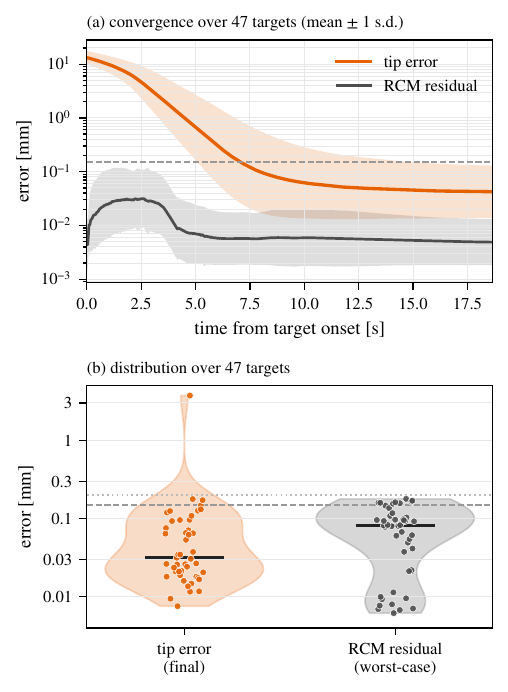}
\caption{\textbf{Hardware test} over $47$ tracked targets at four
RCM entry points. \textbf{(a)} Aggregate convergence of the tip error
and RCM residual, shown as the geometric mean with a $\pm1$ s.d.\
band in log space. \textbf{(b)} Per-target distributions of the final
tip error and worst-case RCM residual; points denote individual
targets and black bars denote medians. Dashed and dotted grey lines
mark the \SI{0.15}{\milli\meter} reach tolerance and
\SI{0.2}{\milli\meter} software-RCM control requirement, respectively.}
\label{fig:hardware}
\end{figure}

Across the four entry points, the unified law ran stably in real time, achieving a pooled reach rate of \SI{94}{\percent} at the \SI{0.15}{\milli\meter} tolerance ($44/47$ targets), a median final tip-regulation error of \SI{32}{\micro\meter} (p95 \SI{164}{\micro\meter}), a median/worst-case kinematically reconstructed RCM residual of \SI{6}{\micro\meter}/\SI{178}{\micro\meter}, and median tip jitter of \SI{22}{\micro\meter}. As shown in Fig.~\ref{fig:hardware}, the aggregate tip error crosses the reach tolerance, while the RCM residual remains roughly an order of magnitude lower; all per-target worst-case RCM residuals remain within the adopted \SI{0.2}{\milli\meter} software-RCM control requirement.


\section{Conclusion}
\label{sec:conclusion}
When a general-purpose manipulator enforces the RCM as a high-priority level of a hierarchical controller, large tip jitter on a small set of reachable targets occurs when the RCM-projected tip Jacobian loses rank while the unprojected tip Jacobian remains well conditioned, a constraint-induced task singularity.
In the observed oscillatory trajectories, this rank loss coincides with rapid rotation of the weak joint-space direction and command reversals.

Because these collapses reflect recoverable configurations rather than unreachable targets, we route around them at the reference level: an authority-guided planner selects a well-conditioned configuration and drives an RCM-consistent reference to it through the low-level QP, embedded in a unified execution law with a single planning gate.

On the collapse case study, the planner reduces tip jitter by more than an order of magnitude. On the fixed $200$-scenario benchmark, the nominal-and-recovery protocol reaches \SI{100}{\percent} of the singularity cases and \SI{99}{\percent} of the near-limit cases while maintaining the software-RCM residual below \SI{0.142}{\milli\meter}. The hardware experiment further demonstrates stable execution of the same architecture at \SI{50}{\hertz}, with the reconstructed RCM residual remaining within \SI{200}{\micro\meter}.
The present planner does not yet enforce anatomical or collision constraints. Adding these constraints and generalising across manipulator kinematics and RCM geometries remains future work.


\bibliographystyle{IEEEtran}
\bibliography{IEEEabrv, Bibliography}
\end{document}